%% file: main.tex
\documentclass{article}

\usepackage{microtype}
\usepackage{graphicx}
\usepackage{subfigure}
\usepackage{booktabs} 
\usepackage{comment}
\usepackage{multirow}
\usepackage{hyperref}
\usepackage[dvipsnames]{xcolor}
\usepackage{xcolor}
\definecolor{Note_color}{rgb}{0.0, 0.0, 1.0}

\usepackage[accepted]{mlsys2023}

\mlsystitlerunning{Towards Cognitive AI Systems: a Survey and Prospective on Neuro-Symbolic AI}

\begin{document}

\twocolumn[
\mlsystitle{Towards Cognitive AI Systems: a Survey and Prospective on Neuro-Symbolic AI \vskip -0.05in}



\mlsyssetsymbol{equal}{*}

\begin{mlsysauthorlist}
\mlsysauthor{Zishen Wan}{to}
\mlsysauthor{Che-Kai Liu}{equal,to}
\mlsysauthor{Hanchen Yang}{equal,to}
\mlsysauthor{Chaojian Li}{equal,to}
\mlsysauthor{Haoran You}{equal,to}
\mlsysauthor{Yonggan Fu}{to}
\mlsysauthor{Cheng Wan}{to}
\mlsysauthor{Tushar Krishna}{to}
\mlsysauthor{Yingyan (Celine) Lin}{to}
\mlsysauthor{Arijit Raychowdhury}{to}
\end{mlsysauthorlist}

\mlsysaffiliation{to}{Georgia Institute of Technology}

\mlsyscorrespondingauthor{Tushar Krishna}{tushar@ece.gatech.edu}
\mlsyscorrespondingauthor{Yingyan (Celine) Lin}{ylin715@gatech.edu}
\mlsyscorrespondingauthor{Arijit Raychowdhury}{arijit.raychowdhury@ece.gatech.edu}

\mlsyskeywords{Machine Learning, MLSys}

\vskip 0.15in

\begin{abstract}
The remarkable advancements in artificial intelligence (AI), primarily driven by deep neural networks, have significantly impacted various aspects of our lives. However, the current challenges surrounding unsustainable computational trajectories, limited robustness, and a lack of explainability call for the development of next-generation AI systems. Neuro-symbolic AI (NSAI) emerges as a promising paradigm, fusing neural, symbolic, and probabilistic approaches to enhance interpretability, robustness, and trustworthiness while facilitating learning from much less data. Recent NSAI systems have demonstrated great potential in collaborative human-AI scenarios with reasoning and cognitive capabilities. In this paper, we provide a systematic review of recent progress in NSAI and analyze the performance characteristics and computational operators of NSAI models. Furthermore, we discuss the challenges and potential future directions of NSAI from both system and architectural perspectives.
\end{abstract}
]



\printAffiliationsAndNotice{\mlsysEqualContribution} 

\input{Sections/1_Introduction}

\input{Sections/2_Survey}
\input{Sections/3_Profile}

\input{Sections/4_Outlook}
\input{Sections/5_Conclusion}
\input{Sections/6_Acknowledgement}

\nocite{langley00}

\bibliography{ref}
\bibliographystyle{mlsys2023}



\end{document}

%% file: Sections/1_Introduction.tex
\section{Introduction}
\vspace{-0.3em}
\label{sec:intro}
Remarkable advancements in artificial intelligence (AI) have had a profound impact on our lives and numerous industries. These advancements are primarily driven by deep neural networks (DNN) and a virtuous cycle involving large networks, extensive datasets, and augmented computing power. As we reap the benefits of this success, there is growing evidence that continuing our current trajectory may not be viable for realizing AI's full potential. 
First, the escalating computational requirements and energy consumption associated with AI are on an unsustainable trajectory~\cite{wu2022sustainable}, threatening to reach a level that could stifle innovation by restricting it to a select few organizations. Second, the lack of robustness and explainability remains a significant challenge, likely due to inherent limitations in current learning methodologies~\cite{wan2021analyzing,dwivedi2023explainable}. Third, contemporary AI systems mostly operate in isolation, with limited collaboration between humans and AI agents. Hence, it is imperative to develop next-generation AI paradigms that address the growing demand for efficiency, explainability, and trust in AI systems.

\begin{figure}[b!]
\vspace{-2em}
        \includegraphics[width=.9\columnwidth]{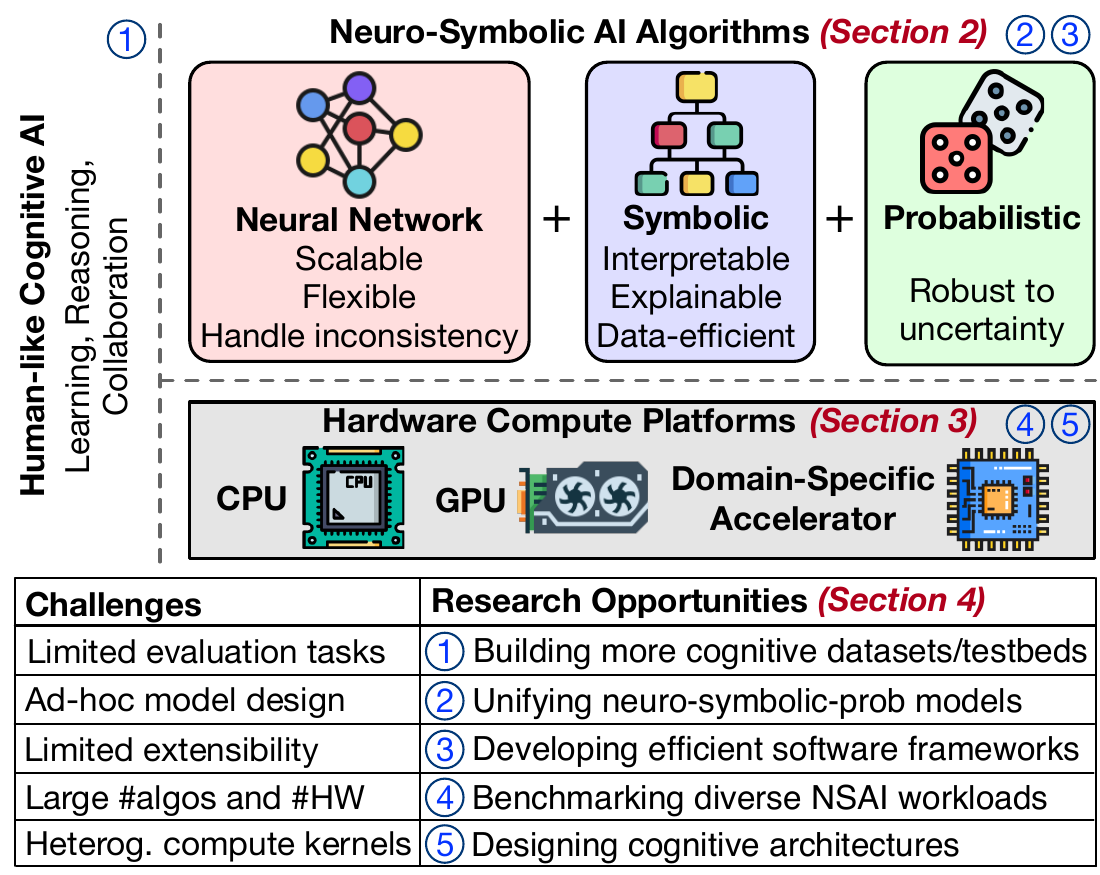}
        \vspace{-1em}
        \centering
         \caption{Overview of neuro-symbolic AI systems, and challenges and opportunities in advancing next-generation cognitive AI.}
        \label{fig:overview}
\end{figure}

Neuro-symbolic AI (NSAI) represents an emerging AI paradigm that integrates neural, symbolic, and probabilistic approaches to enhance explainability, robustness, and enable learning from much less data in AI (Fig.~\ref{fig:overview}). Neural methods have proven highly effective in extracting complex features from data for tasks such as natural language processing and object detection. On the other hand, symbolic methods enhance explainability and reduce the dependence on extensive training data by incorporating established models of the physical world, and probabilistic methods enable cognitive systems to more effectively handle uncertainty, resulting in improved robustness under unstructured conditions. The synergistic fusion of neural, symbolic, and probabilistic methods positions NSAI as a promising paradigm capable of ushering in the third wave of AI~\cite{garcez2023neurosymbolic}.

NSAI promises many possibilities for AI systems that acquire human-like communication and reasoning capabilities, enabling them to recognize, classify, and adapt to new situations autonomously. In addition to its superior performance compared to traditional AI models in tasks such as image and video question answering~\cite{maoneuro,yang2020neurasp}, NSAI holds significant potential for enhancing real-time responses, energy efficiency, explainability, and trustworthiness of collaborative human-AI applications. These applications include collaborative robotics, mixed-reality systems, and human-AI interactions in the metaverse, where robots can seamlessly interact with humans in complex environments, AI agents can reason and make decisions in a robust and explainable manner, and intelligence is pervasively embedded and untethered from the cloud.

In this paper, we provide a systematic survey, evaluation, and analysis of NSAI systems, promising a next-generation AI paradigm.
First, we review and categorize the state-of-the-art NSAI systems from a structured perspective (Sec.~\ref{sec:survey}).
Second, we analyze various NSAI workloads on hardware platforms, examining their runtime characteristics and underlying compute operators (Sec.~\ref{sec:profile}).
Lastly, we discuss the challenges and opportunities for NSAI system research and our outlook on the road ahead (Sec.~\ref{sec:challenge}).
To the best of our knowledge, this is the \emph{first} paper to assess NSAI from both a system and architecture perspective, aiming to inspire the design of next-generation cognitive computing systems through synergistic advancements in NSAI algorithms, systems, architecture, and algorithm-hardware co-design.

%% file: Sections/2_Survey.tex
\vspace{-0.5em}
\section{Neuro-Symbolic AI Algorithms}
\vspace{-0.3em}
\label{sec:survey}
In this section, we systematically review and categorize the recent research progress in NSAI algorithms.
\textbf{Overview.} NSAI represents an interdisciplinary approach that synergistically combines symbolic reasoning with neural network (NN) learning to create intelligent systems, leveraging the complementary strengths of both to enhance the accuracy and interpretability of the resulting models. Given that NSAI algorithms typically incorporate both symbolic and neural components, various NSAI paradigms can be categorized based on how these components are integrated into a cohesive system. Inspired by Henry Kautz's NSAI taxonomy~\cite{henry2020taxonomy}, we systematically categorize these NSAI algorithms into five paradigms, as summarized in Tab.~\ref{tab:summary}. We elaborate each of these paradigms below. Additionally, Tab.~\ref{tab:operation_example} provides examples of several underlying operations based on the categorization in Tab.~\ref{tab:summary}.

\textbf{Symbolic[Neuro]} refers to an intelligent system that empowers symbolic reasoning with the statistical learning 
capabilities of NNs. These systems typically consist of a comprehensive symbolic problem solver that includes loosely-coupled neural subroutines for statistical learning. 
Examples include DeepMind's AlphaGo \cite{silver2017mastering} and AlphaZero~\cite{zhang2020alphazero}, which use Monte-Carlo Tree Search (MCTS) as the symbolic solver and NN state estimators for learning statistical patterns. 

\textbf{Neuro$|$Symbolic} refers to a hybrid system that combines a neural system and a symbolic system in a pipeline, where each component typically specializes in complementary tasks within the pipeline. To the best of our knowledge, the majority of NSAI algorithms fall into this category.
For example, IBM's neuro-vector-symbolic architecture (NVSA) \cite{hersche2023neuro} uses an NN as the frontend for perception and semantic parsing, and a symbolic reasoner as the backend for probabilistic abductive reasoning on the RAVEN~\cite{zhang2019raven} and I-RAVEN~\cite{hu2021stratified} datasets. Other examples include neuro-probabilistic soft logic (NeuPSL) \cite{pryor2022neupsl}, neural probabilistic logic programming (DeepProbLog)~\cite{manhaeve2021neural}, neuro-answer set programming (NeurASP)~\cite{yang2020neurasp}, nerual symbolic dynamic reasoning~\cite{yi2020clevrer}, neural symbolic concept learner (NSCL) \cite{maoneuro}, abductive learning (ABL) \cite{dai2019bridging}, and neuro-symbolic visual question answering (NSVQA) \cite{yi2018neural} on the CLEVRER dataset \cite{yi2020clevrer}.

\textbf{Neuro:Symbolic$\rightarrow$Neuro} approach incorporates symbolic rules into NNs to guide the learning process, where symbolic knowledge is compiled into the structure of neural models for enhancing the model interpretability. 
For instance, logical NNs (LNNs)~\cite{riegel2020logical} encode knowledge or domain expertise as symbolic rules (first-order logic or fuzzy logic) that act as constraints on the NN output. Other examples include deep learning for symbolic mathematics \cite{lampledeep} and differentiable inductive logic programming (ILP) \cite{evans2018learning}.

\input{Tabs/tab_temp}

\textbf{$\mbox{Neuro}_{\mbox{Symbolic}}$} is a type of hybrid approach that combines symbolic logic rules with NNs. It involves mapping symbolic logic rules onto embeddings that serve as soft constraints or regularizers on the NN's loss function.
Logical tensor networks (LTNs) \cite{badreddine2022logic}, for instance, use logical formulas to define constraints on the tensor representations, which has proven successful in knowledge graph completion tasks. These tasks aim to predict missing facts or relationships between entities. Other examples of this approach include deep ontology networks \cite{hohenecker2020ontology} and tensorization methods~\cite{garcez2019neural}. As the inference is still governed by NNs, it remains a research question whether this approach will compromise the interpretability.

\input{Tabs/tab_operation}


\textbf{Neuro[Symbolic]} refers to a system that empowers NNs with the explainability and robustness of symbolic reasoning. Unlike $\textbf{Symbolic[Neuro]}$, where symbolic reasoning is used to guide the neural model learning process, in $\textbf{Neuro[Symbolic]}$, the neural model incorporates symbolic reasoning by paying attention to specific symbolics at certain conditions. For instance, graph neural networks (GNNs) are often adopted as strong candidates for representing symbolic expressions when endowed with attention mechanisms \cite{lamb2020graph}. In particular, this attention mechanism can be leveraged to incorporate symbolic rules into GNN models, enabling selective attention to pertinent  
symbolic information in the graph. Other examples include neural logic machines (NLM)~\cite{dongneural}.





%% file: Tabs/tab_temp.tex
\begin{table}[t!]
\vspace{-7pt}
\centering
\caption{Survey of recent NSAI algorithms into five groups,  with a summary of their underlying operations and vector formats\vspace{3pt}.}
\renewcommand*{\arraystretch}{1.1}
\resizebox{1\columnwidth}{!}{
\begin{tabular}{|c|l|c|c|}
\hline
\textbf{\begin{tabular}[c]{@{}c@{}}Category\end{tabular}}   & \multicolumn{1}{c|}{\textbf{NSAI Algorithm}}                             & \textbf{\begin{tabular}[c]{@{}c@{}} Underlying Operation\end{tabular}} & \multicolumn{1}{c|}{\textbf{If Vector}}                \\ \hline
\textbf{Symbolic{[}Neuro{]}}                                     & \textbf{AlphaGo} \cite{silver2017mastering}     &    NN, MCTS                                                                                  & Vector                 \\ \hline
\multirow{6}{*}{\textbf{Neuro$|$Symbolic}} & \textbf{NVSA}~\cite{hersche2023neuro}  & NN, mul, add       &   Vector \\ \cline{2-4} 
                   & \textbf{NeuPSL} \cite{pryor2022neupsl}   &  NN, fuzzy logic      &              Vector  \\ \cline{2-4} 
                & \textbf{NSCL}  \cite{maoneuro} & NN, add, mul, div, log    &      Vector \\ \cline{2-4} 
                   & \textbf{NeurASP} \cite{yang2020neurasp} &  NN, logic rules     &   Non-Vector \\ \cline{2-4} 
                   & \textbf{ABL}  \cite{dai2019bridging} &NN, logic rules    &       Non-Vector \\ \cline{2-4} 
                   & \textbf{NSVQA}  \cite{yi2018neural}  &  NN, pre-defined objects      &     Non-Vector \\ \hline
\multirow{5}{*}{\begin{tabular}[l]{@{}l@{}}\textbf{Neuro:Symbolic}\\ \textbf{$\rightarrow$Neuro}\end{tabular}} & \textbf{LNN} \cite{riegel2020logical}  &  NN, fuzzy logic               & Vector\\ \cline{2-4} 

                   & \begin{tabular}[l]{@{}l@{}}\textbf{Symbolic Mathematics} \\ \cite{lampledeep}\end{tabular} &  NN                                                                            & Vector \\ \cline{2-4} 
                   
                   & \begin{tabular}[l]{@{}l@{}}\textbf{Differentiable ILP}\\ \cite{evans2018learning}\end{tabular} &  NN, fuzzy logic                                                                                     & Vector \\ \hline
\multirow{3}{*}{\textbf{$\mbox{Neuro}_{\mbox{Symbolic}}$ }} & \textbf{LTN} \cite{badreddine2022logic}  & NN, fuzzy logic                                                                    & Vector \\ \cline{2-4} 
                   & \begin{tabular}[l]{@{}l@{}}\textbf{Deep ontology networks}\\ \cite{hohenecker2020ontology}\end{tabular}  & NN  & Vector  \\ \hline
\multirow{2}{*}{\textbf{Neuro{[}Symbolic{]}} }                  & \begin{tabular}[l]{@{}l@{}}\textbf{GNN+attention}\\ \cite{lamb2020graph}\end{tabular} & NN, SpMM, SDDMM & Vector\\ \cline{2-4} 
& \textbf{NLM} \cite{dongneural}  &  NN, permutation         &  Vector \\  \hline
\end{tabular}}
\label{tab:summary}
\vspace{-1.5em}
\end{table}

%% file: Tabs/tab_operation.tex
\begin{table}[t!]
\vspace{-0.5em}
\centering
\caption{Enumeration of the underlying operations based on Tab.~\ref{tab:summary}\vspace{3pt}.}
\renewcommand*{\arraystretch}{1.1}
\resizebox{1\columnwidth}{!}{
\begin{tabular}{|c|c|}
\hline
\textbf{Underlying Operations}                   & \textbf{Examples}  \\ \hline
\multirow{2}{*}{\begin{tabular}[c]{@{}c@{}}Fuzzy logic\\ (LTN)\end{tabular}} & $F = \forall x (isCarnivor(s)) \rightarrow (isMammal(x))$  \\ 
                    &  $\{isCarnivor(s)$:$[0,1]$, $isMammal(x)$:$[1,0]$\} $\rightarrow F = [1,0]$ \\ \hline
\multirow{1}{*}{Mul and Add (NVSA)}  & $X_i \in \{+1, -1\}^d \rightarrow (X_i \cdot X_j) / (X_i + X_j)$ \\ \hline
\multirow{3}{*}{\begin{tabular}[c]{@{}c@{}}Logic rules\\ (ABL)\end{tabular}} & Domain: $animal (dog). carnivore(dog).mammal(dog)$ \\ 
                    & Logical formula: $mammal(x)\wedge carnivore(x)$ \\
                    & ABL: $hypos(x): -animal(x),mammal(x),carnivore(x)$
                    \\ \hline
\multirow{2}{*}{\begin{tabular}[c]{@{}c@{}}Pre-defined objects\\ (NSVQA)\end{tabular}} & \texttt{equal\_color:} $(entry, entry) \rightarrow Boolean $ \\ 
                    & \texttt{equal\_integer:} $(number, number) \rightarrow Boolean $ \\ \hline
\end{tabular}}
\label{tab:operation_example}
\vspace{-1.5em}
\end{table}

%% file: Sections/3_Profile.tex
\vspace{-0.5em}
\section{NSAI System Profiling}
\vspace{-0.3em}
\label{sec:profile}
In this section, we analyze the performance characteristics of three recent NSAI models and discuss their system bottleneck, workload operators, and optimization opportunities.

\vspace{-0.5em}
\subsection{NSAI Model and Experimental Methodology}
\vspace{-0.3em}
\label{subsec:method}
\textbf{Model Overview.} We select three NSAI models for profiling analysis: an LNN on a logic program task~\cite{riegel2020logical}, an LTN on a binary classification task~\cite{badreddine2022logic}, and an NVSA~\cite{hersche2023neuro} on the Raven's Progressive Matrices task~\cite{zhang2019raven}, representing Neuro:Symbolic$\rightarrow$Neuro, $\mbox{Neuro}_{\mbox{Symbolic}}$, and Neuro$|$Symbolic NSAI systems (Sec.~\ref{sec:survey}), respectively. The readers are referred to their references for more details.

\textbf{Runtime Profiling Method.} We first conduct function-level profiling to capture runtime statistics, and then use the PyTorch Profiler~\cite{pytorch_profiler} to measure the CPU and GPU runtimes of each model at a per-function granularity. The experiments are conducted on a system with an Intel Xeon Silver 4114 CPU and an Nvidia RTX 2080 Ti GPU.

\textbf{Compute Operator Analysis Method.} On top of the above profiling, we perform compute operator-level profiling for further analysis. We classify each neuro and symbolic workload of the LNN, LTN, and NVSA models into six operator categories: convolution, matrix multiplication (MatMul), vector/element-wise operation (e.g., tensor add, div, and norm), data transformation (e.g., reshape and transpose), data movement (e.g., inter-device transfer), and others (e.g., fuzzy logic, and logic rule)~\cite{susskind2021neuro}.

\vspace{-0.5em}
\subsection{NSAI Model Profiling Results}
\label{subsec:results}
\begin{figure}[b!]
\centering
\vspace{-1.7em}
        \includegraphics[width=1.02\columnwidth]{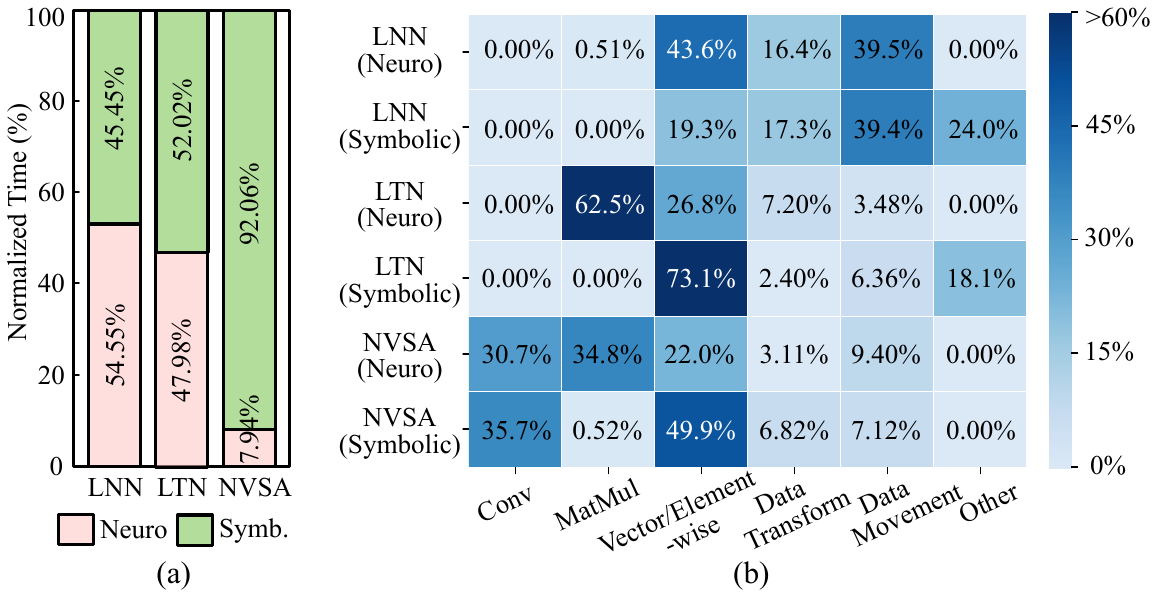}
        \centering
        \vspace{-2em}
        \caption{(a) Runtime breakdown and (b) compute operation analysis for three neuro-symbolic AI models - LNN, LTN and NVSA.}
        \label{fig:operator}
\end{figure}

\vspace{-0.3em}
\textbf{Runtime Breakdown.}
Compared to neuro workloads, symbolic workloads are not negligible in computing latency and may become a system bottleneck. Fig.~\ref{fig:operator}\textcolor{Blue}{(a)} shows the runtime breakdown for each model, where the neuro (symbolic) workloads account for 54.6\% (45.4\%), 48.0\% (52.0\%), 7.9\% (92.1\%) runtime of the LNN, LTN, and NVSA models, respectively. Notably, the symbolic workload dominates the NVSA's runtime, predominately due to the sequential and computational-intensive rule detection during the involved reasoning procedure. This reasoning computation depends on the result of the frontend neuro workload and thus lies on the critical path during inference; Nevertheless, there are still opportunities for leveraging data pre-processing and parallel rule query to reduce this bottleneck.

\textbf{Runtime Scalability.} We observe that the neuro vs. symbolic runtime proportion shown in Fig.~\ref{fig:operator}\textcolor{Blue}{(a)} remains relatively stable across various test sets under the same size, whereas the total runtime increases quadratically with the test set size. For example, when the test set size increases from 2$\times$2 to 3$\times$3, the symbolic workload runtime percentage only increases from 92.06\% to 94.71\%, but the total runtime of the NVSA model increases by 5.02$\times$, indicating the potential scalability bottleneck of NSAI models.

Fig.~\ref{fig:operator}\textcolor{Blue}{(b)} partitions the neuro and symbolic workloads of the LNN, LTN, and NVSA models into six operator categories with runtime breakdown. Below is the workload analysis:

\textbf{Neuro Workload Analysis.}
The neuro workload is dominated by the MatMul and activation operations. LTN (neuro) is dominated by MatMul due to its heavy MLP components, while NVSA's (neuro) majority runtime is on MatMul and convolution because it adopts ResNet18 as the perception backbone for feature extraction. By contrast, a large portion of LNN's (neuro) runtime is on element-wise operations due to the sparse syntax tree structure composed of proposition logic. Notably, data movement also takes up a significant amount of LNN's (neuro) runtime because of its unique bidirectional dataflow during reasoning inference.

\textbf{Symbolic Workload Analysis.}
The symbolic workload is dominated by vector and scalar operations exhibiting low operational intensities and complex control flows. Both LNN (symbolic) and LTN (symbolic) have a large number of logic operations, posing parallelism optimization opportunities in their database queries and arithmetic operations, especially for larger symbolic models. Meanwhile, LNN (symbolic) is severally data movement-bounded due to its sparse and irregular memory accesses and bidirectional inference, where model-aware dataflow architecture would likely be beneficial for alleviating this bottleneck.
Notably, the element-wise operations usually stem from high-dimensional distributed vector computations (e.g., binding, bundling, and permutation) for symbolic representation, which is difficult to process efficiently on GPUs.
Therefore, the challenges of accelerating these computations will become increasingly important as the task and feature complexities further grow.


%% file: Sections/4_Outlook.tex
\vspace{-0.5em}
\section{Challenges and Opportunities}
\vspace{-0.3em}
\label{sec:challenge}
In this section, we discuss the challenges and opportunities for NSAI systems, and outlines our vision for the future, focusing on the system and architecture perspectives.

\textbf{Building ImageNet-like NSAI datasets.}
NSAI systems hold great potential in achieving human-like AI~\cite{booch2021thinking}. However, their current applications are limited to basic decision-making and reasoning problems~\cite{garcez2022neural}, falling short of the broader vision of human cognitive abilities, such as interpretability, deductive reasoning, systematicity, productivity, compositionality, inferential coherence of mental thought, and causal and counterfactual thinking. To significantly advance the metacognitive capabilities of NSAI systems, more challenging and suitable datasets are highly desirable to unleash NSAI's potential.

\textbf{Unifying neuro-symbolic-probabilistic models.}
Integrating neural, symbolic, and probabilistic approaches offers promise to improve AI models' explainability and robustness. However, the current attempts to combine these complementary approaches are still in a nascent manner~\cite{wang2022towards} - how to integrate them in a principled manner remains a fundamental and open challenge. We envision a unified framework to design algorithms that opportunistically combine neural, symbolic, and probabilistic components, and for quantifying scaling laws for neuro-probabilistic-symbolic inference versus large neural models. 

\textbf{Developing efficient software frameworks.}
NSAI systems typically utilize underlying logic, such as fuzzy logic, parameterization, and differentiable structures, to support learning and reasoning capabilities. 
However, most NSAI system implementations create custom software for deduction for the particular logic used, which limits modularity and extensibility~\cite{aditya2023pyreason}. 
Therefore, new software frameworks are needed that can encompass a broad set of reasoning logical capabilities and provide practical syntactic and semantic extensions while being fast and memory-efficient. Moreover, new programming models, compilers, and runtimes that can facilitate the ease and efficient realization of the neuro-symbolic-probabilistic models are of significance to realize the full promise of NSAI paradigms.

\textbf{Benchmarking diverse NSAI workloads.} 
Given the proliferation of NSAI algorithms and the rapid advancements in hardware platforms, it is crucial to benchmark NSAI systems in a comparable, quantitative, and validatable manner.
To achieve this, from the system aspect, we need representative benchmarks that capture the essential workload characteristics (e.g., compute kernels, access patterns, and sparsity) of neural, symbolic, and probabilistic models, and that can be quantitatively tested in human-AI applications.
Additionally, from an architectural and hardware perspective, we need modeling-simulation-characterization frameworks to enable the development of novel architectures for these workloads and build optimized modular blocks as libraries by leveraging workload characteristics, as other emerging domains~\cite{krishnan2022roofline,wan2022robotic}.
Benchmarking NSAI will guide ML researchers and system architects in investigating the trade-offs in accuracy, performance, and efficiency of various NSAI algorithms, and in implementing systems in a performance-portable way.

%

\textbf{Designing cognitive hardware architectures.}
NSAI workloads that combine neural, symbolic, and probabilistic methods feature much greater heterogeneity in compute kernels, sparsity, irregularity in access patterns, and higher memory intensity than current DNN workloads. This leads to an increasing divergence with the current hardware roadmap that largely focuses on matrix multiplication or nearest neighbor search, and regular dataflows, e.g., systolic arrays~\cite{krishnan2022automatic} or compute-in-memory crossbars~\cite{crafton2022improving}. Therefore, we need novel architectures with dedicated processing units, memory hierarchies, and on-chip interconnects that can handle the additional complexities in computations and communications. Additionally, the architecture needs to provide flexibility with both configurable interconnects and full addressable memories to keep pace with NSAI algorithmic innovations.

%% file: Sections/5_Conclusion.tex
\section{Conclusion}
\vspace{-0.3em}
\label{sec:conclusion}
NSAI is an emerging paradigm for next-generation efficient, robust, explainable, and cognitive AI systems. This paper systematically reviews recent NSAI algorithms, characterizes their system performance, analyzes their workload operators, and identifies the challenges and opportunities towards fullfiling next-generation NSAI systems.

%% file: Sections/6_Acknowledgement.tex
\vspace{-5pt}
\section*{Acknowledgements}
We thank Sixu Li and Yang (Katie) Zhao for their technical support. This work was supported in part by CoCoSys, one of the seven centers in JUMP 2.0, a Semiconductor Research Corporation (SRC) program sponsored by DARPA.